\documentclass[letterpaper]{article} 
\usepackage{aaai25}  
\usepackage{times}  
\usepackage{helvet}  
\usepackage{courier}  
\usepackage[hyphens]{url}  
\usepackage{graphicx} 
\usepackage{natbib}  
\usepackage{caption} 
\usepackage{algorithm}
\usepackage{algorithmic}
\usepackage{amsmath}
\usepackage{amssymb}
\usepackage{newfloat}
\usepackage{listings}
\usepackage{booktabs}
\usepackage{tabularx}
\usepackage{amsmath}
\DeclareCaptionStyle{ruled}{labelfont=normalfont,labelsep=colon,strut=off} 
\floatstyle{ruled}
\newfloat{listing}{tb}{lst}{}
\floatname{listing}{Listing}
\title{CAD-GPT: Synthesising CAD Construction Sequence with Spatial Reasoning-Enhanced Multimodal LLMs}
\author{
    Siyu Wang\textsuperscript{\rm 1,2}, 
    Cailian Chen\textsuperscript{\rm 1,2,3}\thanks{Corresponding author}, Xinyi Le\textsuperscript{\rm 1,2}, Qimin Xu\textsuperscript{\rm 1,2}, Lei Xu\textsuperscript{\rm 4,5}, Yanzhou Zhang\textsuperscript{\rm 1,2}, Jie Yang\textsuperscript{\rm 6}
}
\affiliations{
    \textsuperscript{\rm 1}School of Electronics Information and Electrical Engineering, Shanghai Jiao Tong University, Shanghai, China\\
    \textsuperscript{\rm 2}Key Laboratory of System Control and Information Processing, Ministry of Education of China, Shanghai, China\\
    \textsuperscript{\rm 3}SJTU-Paris Elite Institute of Technology, Shanghai Jiao Tong University, Shanghai, China\\
    \textsuperscript{\rm 4}Institute of Cyber Science and Technology, Shanghai Jiao Tong University, Shanghai, China\\
    \textsuperscript{\rm 5}Shanghai Key Laboratory of Integrated Administration Technologies for Information Security, Shanghai, China\\
    \textsuperscript{\rm 6}University of Minnesota Twin Cities, Saint Paul, MN, USA\\

    \{y\_wsy09, cailianchen, lexinyi, qiminxu, xulei1, sjtu\_zyz\}@sjtu.edu.cn, jieyang@umn.edu
}

\usepackage{bibentry}

\begin{document}

\maketitle

\begin{abstract}
Computer-aided design (CAD) significantly enhances the efficiency, accuracy, and innovation of design processes by enabling precise 2D and 3D modeling, extensive analysis, and optimization. Existing methods for creating CAD models rely on latent vectors or point clouds, which are difficult to obtain and costly to store. Recent advances in Multimodal Large Language Models (MLLMs) have inspired researchers to use natural language instructions and images for CAD model construction. However, these models still struggle with inferring accurate 3D spatial location and orientation, leading to inaccuracies in determining the spatial 3D starting points and extrusion directions for constructing geometries. This work introduces CAD-GPT, a CAD synthesis method with spatial reasoning-enhanced MLLM that takes either a single image or a textual description as input. To achieve precise spatial inference, our approach introduces a 3D Modeling Spatial Mechanism. This method maps 3D spatial positions and 3D sketch plane rotation angles into a 1D linguistic feature space using a specialized spatial unfolding mechanism, while discretizing 2D sketch coordinates into an appropriate planar space to enable precise determination of spatial starting position, sketch orientation, and 2D sketch coordinate translations. Extensive experiments demonstrate that CAD-GPT consistently outperforms existing state-of-the-art methods in CAD model synthesis, both quantitatively and qualitatively. 
\begin{links}
    \link{Project Page}{https://OpenIWIN.github.io/CAD-GPT/}
\end{links}
\end{abstract}

%

\section{Introduction}

\begin{figure*}[htbp]
    \centering  
    \includegraphics[width=\linewidth]{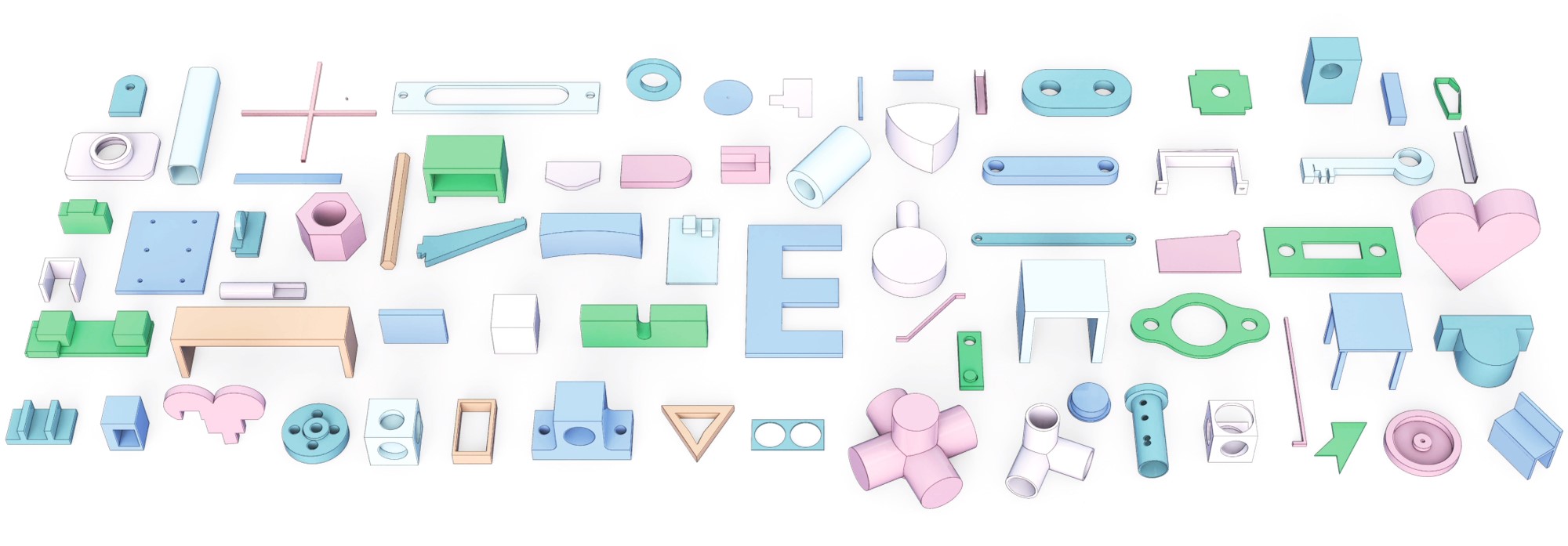} 
    \caption{Demonstration of various CAD models generated by CAD-GPT. The models in the image demonstrate semantic sketch generation capabilities (e.g., a heart shape and the letter "E"), category-based CAD generation capabilities (e.g., a table, a chair, and a key), spatial reasoning abilities (e.g., a table and mutually perpendicular cylinders), and the capability to generate identical models with varying dimensions (e.g., three connectors with two circular holes of differing sizes).}
    \label{fig:wide_image}
\end{figure*}
Computer-Aided Design (CAD) has become the standard approach for designing, drafting, and modeling in a wide range of industries\cite{robertson1993cad,chen2024sets}. Almost every manufactured object that exists today started its life in a parametric CAD tool. The CAD command sequence is one type of CAD model representation. It is described as a sequence of operations such as drawing 2d sketches and extruding sketches into 3D solid shapes\cite{wu2021deepcad}. Constructing these CAD models requires domain expertise and spatial inference capabilities, and it can also be time-consuming. 

Recently, the most popular direction for CAD model generation focused on using generative models like variational autoencoder(VAE) (Wu, Xiao, and Zheng 2021) and vector quantized variational autoencoder(VQ-VAE) (Xu et al. 2022, 2023). These methods map CAD models to vectors or codebooks in a high-dimensional latent space and then reconstruct the original CAD models from these high-dimensional representations. The main limitations of both methods include: 1) The quality of CAD models synthesized by these methods depends not only on the methods’ capabilities but also on the quality of the provided guidance vectors or codebooks, which can inevitably result in cumulative errors. 2) These methods require high dimensional data, similar to the distribution of their vectors or codebooks as inputs, which are difficult to obtain directly. Another line of work directly infers CAD sequences from point clouds(Ma et al. 2023; Khan et al. 2024) or sketches(Li et al. 2022). In practical applications, sketches need to be drawn by professionals, and point clouds require specialized equipment for collection, both of which involve high data acquisition costs.

Generative AI tools such as Multimodal Large Language Models (MLLMs) have the potential to remove these barriers. These multimodal models exhibit impressive visual language understanding and generation capabilities\cite{achiam2023gpt, yin2023survey}. Recently, there have been initial attempts to use state-of-the-art MLLMs for the creation of CAD models\cite{makatura2023can, badagabettu2024query2cad}. Experiments show that these models, such as GPT-4, lack spatial reasoning capabilities\cite{makatura2023can} and have a low success rate\cite{badagabettu2024query2cad} in generating the desired CAD models. These limitations can manifest as notable challenges in the design and manufacturing domain. For instance, they may generate a car with four horizontally placed wheels or a table with legs that exceed the tabletop and are randomly positioned. Hence, the main question we ask is: \textit{\textbf{How to enhance the 3D spatial reasoning capabilities of multimodal large language models for accurate CAD model synthesis?}}

In this paper, we introduce CAD-GPT, a MLLM with enhanced 3D spatial reasoning capability built upon LLaVA-1.5 7B version\cite{liu2024improved}. For training the model, we constructed a dataset that pairs CAD modeling sequences with natural language descriptions and single fixed-view rendered images of the CAD models. We built our dataset based on the DeepCAD dataset\cite{wu2021deepcad}. To enhance the spatial reasoning capabilities of the model, we developed a 3D spatial localization mechanism specifically tailored for 3D modeling tasks. Concretely, we convert the global spatial 3D coordinates, sketch plane rotation angles into two distinct categories of position tokens by unfolding their characteristics into a 1D linguistic feature space. Additionally, the 2D sketches are discretized and converted into special tokens. These tokens are incorporated into the vocabulary of the base LLM. Simultaneously, we incorporate custom learnable positional embeddings to bridge the gap between language and spatial positions.

In summary, our contributions are as follows:

\begin{itemize}   
    \item We present CAD-GPT, a MLLM that synthesises CAD modeling sequences precisely from a single image or textual description. To the best of our knowledge, we are the first to develop a MLLM specifically trained for this task.
    \item  We designed a novel localization mechanism tailored for the 3D modeling process, enhancing the spatial reasoning capabilities of large-language models by mapping 3D space into 1D through a tokenization method.
    \item Utilizing the DeepCAD dataset, we generated 160k fixed-viewpoint CAD model images and 18k corresponding natural language captions. We plan to release our CAD-GPT model along with the dataset we developed, contributing a valuable resource.
    \item Experiments on the held-out dataset demonstrate that our approach achieves a higher accuracy compared to state-of-the-art baseline models.
\end{itemize}

\begin{figure*}  
    \centering   
    \includegraphics[width=\linewidth]{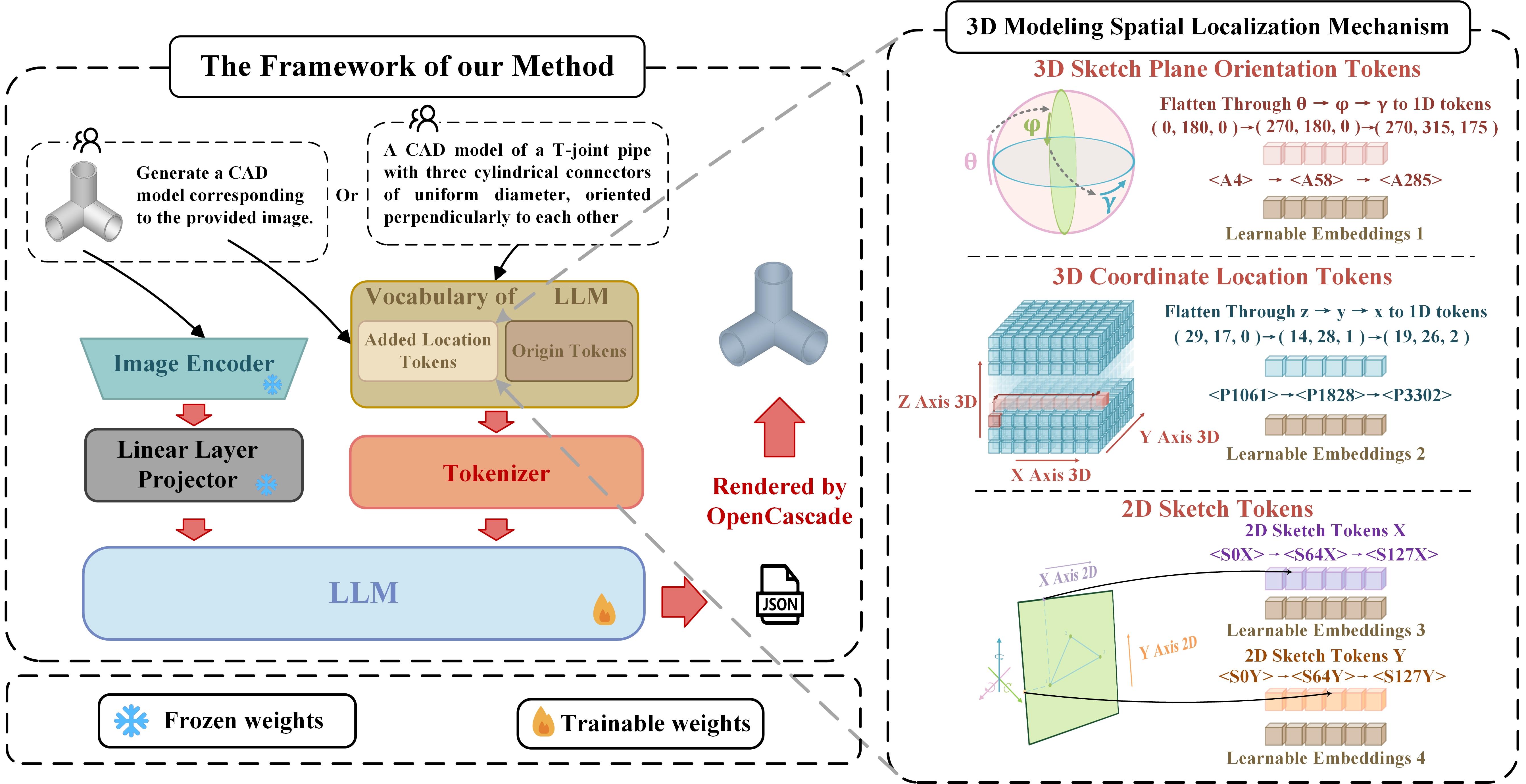}
    \caption{Overview of our CAD-GPT framework. On the left side, a dashed box contains the overall algorithm framework. The right side provides a detailed view of our 3D Modeling Spatial Localization Mechanism. From top to bottom, it sequentially demonstrates the 3D Orientation, 3D Coordinate Location, showing how they unfold from 3D to 1D along specific directions, as well as the token representation method for the 2D Sketch.}
    \label{fig:wide_image}
\end{figure*}

\section{Related Work}

\subsection{Approximate 3D Representation} Accurate and efficient 3D data representation remains a challenge in computer graphics and vision. \textbf{Point Clouds} \cite{zhou20213d, luo2021diffusion, nichol2022point} capture discrete spatial points, offering simplicity but lacking surface details; \textbf{Meshes}\cite{groueix2018papier, wang2018pixel2mesh, chen2024meshanything, siddiqui2024meshgpt} use vertices and edges to form polygons, providing connectivity but facing complexity issues; \textbf{3D Gaussians model} \cite{kerbl20233d, tang2023dreamgaussian} use points with Gaussian distributions for efficient rendering, yet they miss precise surface features; and \textbf{Neural Radiance Fields (NeRF)} \cite{mildenhall2021nerf} employ neural networks for volumetric modeling, requiring substantial computational resources and data. These representations often struggle with noise, incomplete details, and limited editability.
\subsection{Computer-Aided Design Model Representations}
\subsubsection{Direct B-rep Generation}  involves synthesizing the underlying parametric curves and surfaces and the topology that connects them to create a solid model\cite{wang2020pie,sharma2020parsenet}. This work focuses on developing a generative model for CAD construction sequences rather than B-reps. However, converting a B-rep into a construction sequence is challenging, as multiple command sequences can produce the same B-rep.
\subsubsection{CAD Construction Sequence Generation}
DeepCAD \cite{wu2021deepcad} was the first to propose a sketch-extrusion construction sequence representation for CAD models, predicting CAD history from latent vectors or point clouds as a preliminary experiment. HNC-CAD \cite{xu2023hierarchical} introduced a hierarchical code tree representation for CAD sequences based on VQ-VAE, which can autoregressively generate various CAD sequences from different codebooks. However, these codebooks' high-dimensional and abstract nature makes it difficult to generate desired CAD models directly. The recently proposed CAD-SIGNet \cite{khan2024cad} can generate CAD sequences from point clouds and produce CAD models by selecting different sketches from the autoregressively generated intermediates. However, obtaining point clouds requires a time-consuming and costly process with expensive equipment. In the fields of 3D mesh or 3D Gaussians generation, recent work \cite{chen2024meshxl, wang2024gaussianeditor} has leveraged large language models(LLMs) and autoregressive methods to achieve text-to-3D and image-to-3D generation. However, these approaches have yet to be explored in the CAD construction sequence generation domain.
\subsubsection{MLLMs}
Recent advancements in LLMs \cite{mattas2023chatgpt, zhao2023survey} have revealed extraordinary emergent abilities through scaling data and model sizes. Meanwhile, large vision models (LVMs) \cite{kirillov2023segment} excel in visual clarity but often need help with reasoning. Combining these strengths, the emerging field of MLLMs\cite{achiam2023gpt, yin2023survey} integrates LLMs with billion-scale parameters and new training paradigms, such as multimodal instruction tuning. This integration enables MLLMs to generate website code from images, interpret memes, and perform OCR-free math reasoning. The remarkable success of these applications inspires us to extend such methods to CAD construction sequence generation, which can be considered as a form of 3D modeling code generation.
\subsubsection{User-Controlled 2D/3D Modeling Tasks}
Recently, IconShop\cite{wu2023iconshop} has demonstrated the capability of language models to generate SVG 2D vector graphics from text prompts, representing a significant advancement in 2D modeling. Experimental evidence \cite{makatura2023can} indicates that GPT-4 struggles with specific types of reasoning, particularly those requiring analytical and spatial skills. Recently, CAD-Llama \cite{li2025cad} has made a promising attempt at leveraging open-source LLMs to generate parametric CAD models. Query2CAD has developed a pipeline that utilizes GPT-4 and GPT-3.5 to generate CAD modeling code from text descriptions. However, the failure rate has reached  30\%  to 50\%.  Currently, there is no work addressing these issues or extending MLLMs to the domain of 3D modeling or CAD construction sequence generation, nor are there MLLMs specifically fine-tuned for these types of problems.

\section{Method}
\subsection{Overview}
In this section, we first briefly introduce the model architecture of CAD-GPT. After that, we describe the representation of CAD command sequences. Next, we propose the 3D modeling spatial localization mechanism that enhances the spatial reasoning capabilities of the base MLLM.
\subsection{Model Architecture}
An efficient MLLM can be divided into three main modules: a visual encoder \(g\) tasked with processing visual inputs, a pre-trained language model $f_{\phi } \left ( \cdot  \right ) $ parameterized by $\phi$ that manages the received multimodal signals and performs reasoning, and a visual-language projector \(P\) which functions as a bridge to align the two modalities. 
We adopt LLaVA-1.5 7B version \cite{liu2024improved} as our base model with the pre-trained Vicuna \cite{chiang2023vicuna} as our pedestal LLM. Vicuna is built on LLaMA-2 \cite{touvron2023llama}.

For an input image $I_{V}$, utilizing the pre-trained visual encoder ViT-L/14-336px as \(g\), which provides the visual feature  $Z_{v}=g(I_{V})$. We consider a simple two-layer linear layer as the vision-language projector to map the visual patch embeddings $Z_{v}$  into the text feature space: $S_{v} = P(Z_{v})$. Thus, we have a sequence of visual tokens $S_{v}$, which can be understood just like the text tokens $S_{q}$ by the LLM.
Specifically, for a sequence of length \(L\), we compute the probability of the target answers $S_{a}$ by:
\begin{equation}
\small
p(S_{a} \mid I_V, S_{instruct}) = \prod_{i=1}^{L} p_{\Theta}(x_i \mid S_V, S_{instruct}, S_{a}, _{<i})
\end{equation}
where $\Theta$ is the trainable parameters, $S_{instruct}, _{<i}$ and $S_{a}, _{<i}$ are the instruction and answer tokens before the current prediction token $x_i$. 
\subsection{CAD Command Sequence Representation}
Following the DeepCAD dataset \cite{wu2021deepcad}, a CAD model is represented as a sequence of modeling operations that the user executes to construct a 3D   3 d shape. This type of CAD model is saved in JSON format, storing key modeling commands and parameters (see Table 1) in the order of CAD construction. The sequence of commands is human-readable and easily editable. Moreover, JSON is one of the formats used for the LLaMA-2 pre-training corpus and aligns with its prior knowledge. Consequently, we directly preserve the JSON-formatted CAD modeling sequence as the output format of our model.

These commands describe a CAD model \(M\) as a sequence of pairs of curve and extrusion commands interleaved.  In other words, \( M \) is a command sequence \( M = [C_1, \ldots, C_{N_c}] \), where each \( C_i \) has the form \( (t_i, \mathbf{p}_i) \), specifying the command type \( t_i \) and parameters \(\mathbf{p}_i\).
The commands include details for determining the global starting point of the sketch, the angles between the sketch plane and the three coordinate axes, various parameters for drawing the 2D sketch. Based on these commands, the 2D sketches can be iteratively drawn and extruded to form a 3D model.

To execute an extrusion command, one must first define the profile's sketch plane's 3D orientation and spatial location.This ensures that closed curves can be accurately drawn at the correct 2D local starting point on the correctly oriented sketch plane. The orientation of the sketch plane is defined by the parameters \( \theta, \gamma, \phi \). At the same time, the spatial location is specified by the coordinates \( p_x, p_y, p_z \), which denote the origin of the sketch plane (see Table 1).
Sketch commands define closed curves on a 2D plane, with curve parameters specifying the curve's 2D location in the sketch plane's local frame. We consider three widely used curve commands: drawing a line, an arc, and a circle. Precise command types and 2D coordinates are crucial for accurately sketching the design. In summary, a sketch profile \( S \) is described by a list of loops \( S = [Q_1, \ldots, Q_N] \), where each loop \( Q_i \) consists of a series of curve commands, such that \( Q_i = [C_1, \ldots, C_{n_i}] \). Each curve command \( C_j = (t_j, p_j) \) specifies the curve type \( t_j \) and its shape parameters \( p_j \).

\begin{table}
\label{table:1}
\centering

\begin{tabularx}{\columnwidth}{@{} X p{6.0cm} @{}}
\toprule
\multicolumn{1}{c}{\textbf{Commands}} & \multicolumn{1}{c}{\textbf{Parameters}}   \\
\toprule
\multicolumn{1}{c}{Line} & \(x, y\): 2D line start-point\\
     & \(x, y\): 2D line end-point\\
\midrule
\multicolumn{1}{c}{Arc}  & \(x, y\): 2D arc start-point\\
     & \(x, y\): 2D arc mid-point\\
     & \(x, y\): 2D arc end-point\\
\midrule
\multicolumn{1}{c}{Circle} & \(x, y\): 2D circle start-point\\
       & \(x, y\): 2D circle center\\
\midrule
\multicolumn{1}{c}{Extrude} & \begin{tabular}[c]{@{}l@{}}\(\theta, \phi , \gamma\): 3D sketch plane orientation\\ \(p_{x},  p_{y},  p_{z}\) : 3D sketch plane origin\\ \(s\): scale of associated sketch profile\\ \(e_{1},  e_{2}\): extrude distances toward both sides\\ \(b\): boolean type, \(u\): extrude type\end{tabular}\\
\bottomrule
\end{tabularx}
\caption{Origin CAD command sequences and the main parameters.}
\end{table}

\subsection{3D Modeling Spatial Localization Mechanism} 

Selecting the coordinates of the 3D sketch plane origin coordinate, determining the 3D sketch plane orientation, and then drawing the sketch involves complex mathematical and 3D geometric reasoning processes. Our preliminary experiments demonstrate that MLLMs struggle to infer these parameters accurately, leading to low precision and high failure rates in CAD model generation. To address these challenges, we propose a 3D Modeling Spatial Localization Mechanism. 

Specifically, we have designed three series of localization tokens to replace the parameters for sketch plane origin coordinates, orientation angles of the sketch plane, and 2D sketch curve coordinates. These tokens have been added to the LLM's vocabulary, enabling the model to reason about 3D spatial transformations as seamlessly as it generates words. A detailed explanation of this method is provided in the following sections. Each type of token is enclosed by two distinct boundary tokens, which are composed of special tokens. These boundary tokens serve to signal the model to output the corresponding series of localization tokens. \textbf{All tokens constructed for the 3D Modeling Spatial Localization Mechanism are presented in Table 2.}

\begin{table}
\label{table:2}
\centering

\begin{tabularx}{\columnwidth}{@{} X p{3.8cm} @{}}
\toprule
\multicolumn{1}{c}{$\textbf{Commands}$} & \multicolumn{1}{c}{$\textbf{Parameters}$}   \\
\toprule
\multicolumn{1}{c}{3D Orientation Tokens} & 
\multicolumn{1}{c}{$\texttt{<An>}$, \quad $n \in \mathbb{N}_{0}^{728}$}\\
\midrule
\multicolumn{1}{c}{3D Localization Tokens}  & \multicolumn{1}{c}{$\texttt{<Pk>}$, $\quad k \in \mathbb{N}_{0}^{K^3-1}$}\\
\midrule
\multicolumn{1}{c}{2D Sketch Tokens}  &\multicolumn{1}{c}{$\texttt{<SlX>}$, $\texttt{<SmY>}$, $\quad l,m \in \mathbb{N}_{0}^{127}$}\\
\midrule
\multicolumn{1}{c}{Boundary tokens} & \begin{tabular}[c]{@{}l@{}}$\texttt{<angles>}$, $\texttt{</angles>};$\\$\texttt{<spatial\_position>}$, \\ $\texttt{</spatial\_position>}$;\\$\texttt{<sketch\_position\_x>}$ , \\ $\texttt{</sketch\_position\_x>}$;\\$\texttt{<sketch\_position\_y>}$, \\ $\texttt{</sketch\_position\_y>}$;\end{tabular}\\
\bottomrule 

\end{tabularx}
\caption{Tokens customized for the 3D Modeling Spatial Localization Mechanism}
\end{table}

\subsubsection{3D Sketch Plane Orientation Tokens} 

In the CAD construction sequence, the orientation is represented by a rotation matrix composed of three consecutive parameters: \( \theta, \phi, \) and \( \gamma \). This matrix is designed to align the world frame of reference to the plane’s local frame of reference, specifically orienting the \( z \)-axis to match the plane’s normal direction. Following the order of \( \theta \rightarrow \phi \rightarrow \gamma \), progressing from lowest to highest, each angle is discretized into 9 integer values, resulting in a total of 729 orientation tokens represented as 
\begin{equation}
\texttt{<An>}, \quad where\quad n \in \mathbb{N}_{0}^{728}.
\end{equation}
After this, the angles are aligned with the language space of large language models as part of the linguistic structure. 

\subsubsection{3D Coordinate Localization Tokens} 

We normalize each CAD model within a \(1 \times 1 \times 1\) cube. Next, we discretize the vertices' coordinates into \(K^3\) grids, where \(K = 36\). The grids are sorted in \(z \rightarrow y \rightarrow x\) order, from lower to higher, following MeshGPT \cite{siddiqui2024meshgpt} and Polygen \cite{nash2020polygen}. The order indices of these grids are used to construct our position tokens, forming a sequence of location tokens
\begin{equation}
\texttt{<Pk>}, \quad where \quad k \in \mathbb{N}_{0}^{K^3-1}.
\end{equation}
For instance, a spatial point \(O_i\), if its normalized coordinates \((p_{x_i}, p_{y_i}, p_{z_i})\) are located within grid \(o_i\), then its corresponding position coordinate is \(P_{o_i}\).

\subsubsection{2D Sketch Coordinate Tokens} 

After normalizing the 3D model to a \(1 \times 1 \times 1\) cube, we also normalize each 2D sketch profile within its bounding box and quantize their values into 128 levels. Consequently, the \(x\) and \(y\) coordinates are represented by two series of tokens as follows:

\begin{equation}
\left\{
\begin{array}{l}
\texttt{<SlX>} \\
\texttt{<SmY>}
\end{array}
\right.
\quad \text{where} \quad l, m \in \mathbb{N}_{0}^{127}.
\end{equation}
These tokens indicate the discretized levels of \(x\) and \(y\) coordinates for the 2D sketch.

\subsubsection{Augmenting Spatial Features with Position Embeddings} 

We introduced four distinct types of tokens to the vocabulary. Correspondingly, we expanded the embedding layers to accommodate these additional tokens and incorporated learnable position embedding layers to enhance the representation of the four types of spatial information.
The use of learnable position embeddings allows the model to understand the relative positioning and relationships within the spatial data, enhancing the accuracy and expressiveness of the representation.
Specifically, we introduced the following learnable position embedding matrices: \( W_{\text{angle}} \in \mathbb{R}^{729 \times D} \), \( W_{\text{3D\_pos}} \in \mathbb{R}^{K^3 \times D} \), \( W_{\text{2D\_sketch\_x}} \in \mathbb{R}^{128 \times D} \), and \( W_{\text{2D\_sketch\_y}} \in \mathbb{R}^{128 \times D} \). These matrices were used to augment the embeddings of the corresponding token types, enhancing their spatial information representations.

\section{Dataset Construction}
Our work is based on the DeepCAD dataset \cite{wu2021deepcad}, which contains 178,238 CAD modeling sequences. Referring to SkexGen \cite{xu2022skexgen}, we remove duplicate models.

Considering the unique characteristics of CAD models, the division of CAD data in our dataset differs from that of other 3D models. Many models within the dataset are randomly constructed parts without known categories, while others can be succinctly described with a single sentence that encapsulates their category and characteristics. Consequently, our dataset includes fewer instances of natural language descriptions compared to image-based data.

First, using the OpenCascade \cite{opencascade}, we render 2D images for each CAD model from fixed angles. Then we have developed ten distinct natural language modeling instructions designed to guide CAD-GPT in generating CAD models based on reference images. For instance, one of the instructions is: "Please create a CAD model based on the provided image." During each iteration of fine-tuning, a different instruction is randomly selected. Additional instructions are detailed in the supplementary material. 
Ultimately, the dataset for fine-tuning CAD generation from images comprises \(162k\) samples.

In order to generate accurate textual model caption data, We first use GPT-4o to classify and filter the CAD models by combining JSON and rendered images, removing those that cannot be described. This process leaves us with \( 19k \) models.
We established a pipeline for generating instructions based on InstructGPT \cite{wang2022self}, which was used to produce natural language descriptions for a dataset of \(19k \) instances. Following this, we manually curated the generated descriptions to eliminate those deemed irrelevant or erroneous. Consequently, a refined dataset comprising \(18k\) natural language descriptions of CAD models was retained. These curated data was saved in the format specified for fine-tuning LLaVA, excluding image data, and was subsequently utilized for mixed training purposes.

\section{Experiments Settings}

In this section, we first introduce the detailed training parameters and strategies of our method. We then present the CAD generation results for both image and text input conditions. Additionally, we conduct ablation studies to compare the performance of the baseline multimodal model with and without our 3D Modeling Spatial Localization Mechanism, demonstrating the effectiveness of our approach.

\subsection{Implementation Details}

We freeze the linear mapping layer and vision encoder weights of LLaVA, while fully fine-tuning the language base model. We constructed our data input model based on the LLaVA fine-tuning format, incorporating mixed image-CAD sequence data and text-only description-CAD sequence data. The training involves two stages: first training on the image2CAD task, followed by fine-tuning on the text2CAD task with a reduced learning rate. During training, the newly introduced embedding layers are initialized based on the original vocabulary embedding. The network was trained using a batch size of 8 per GPU across \(4  \times  \) NVIDIA RTX A800 GPUs, with a total training duration of 96 hours. The initial learning rate is set to \(2 \times 10^{-5} \), with a Cosine-Warmup learning rate initialization strategy and a warm-up ratio of \(0.3\). Additionally, following an extrapolation optimization strategy, we adjust certain parameters, expanding the model's maximum input sequence length to 8192.
\subsection{Metrics}
To comprehensively evaluate the predicted sequences, we employ a set of metrics that assess different aspects of the predictions. Specifically, the final CAD reconstructions are quantitatively analyzed against ground-truth CAD models using Chamfer Distances (CD) \cite{fan2017point}. Since CAD sequences are predicted as tokens, they may not always generate successfully rendered CAD models when reconstructed with OpenCascade, we introduce an Invalidity Ratio(IR) metric, expressed as a percentage, which quantifies the proportion of invalid models. In addition, we evaluate command accuracy using two metrics: Command Accuracy (\(\text{ACC}_{\text{cmd}}\)) and Parameter Accuracy (\(\text{ACC}_{\text{param}}\)).

\subsection{CAD Generation from a Single Image}
\begin{figure}
    \centering
    \includegraphics[width=1\linewidth]{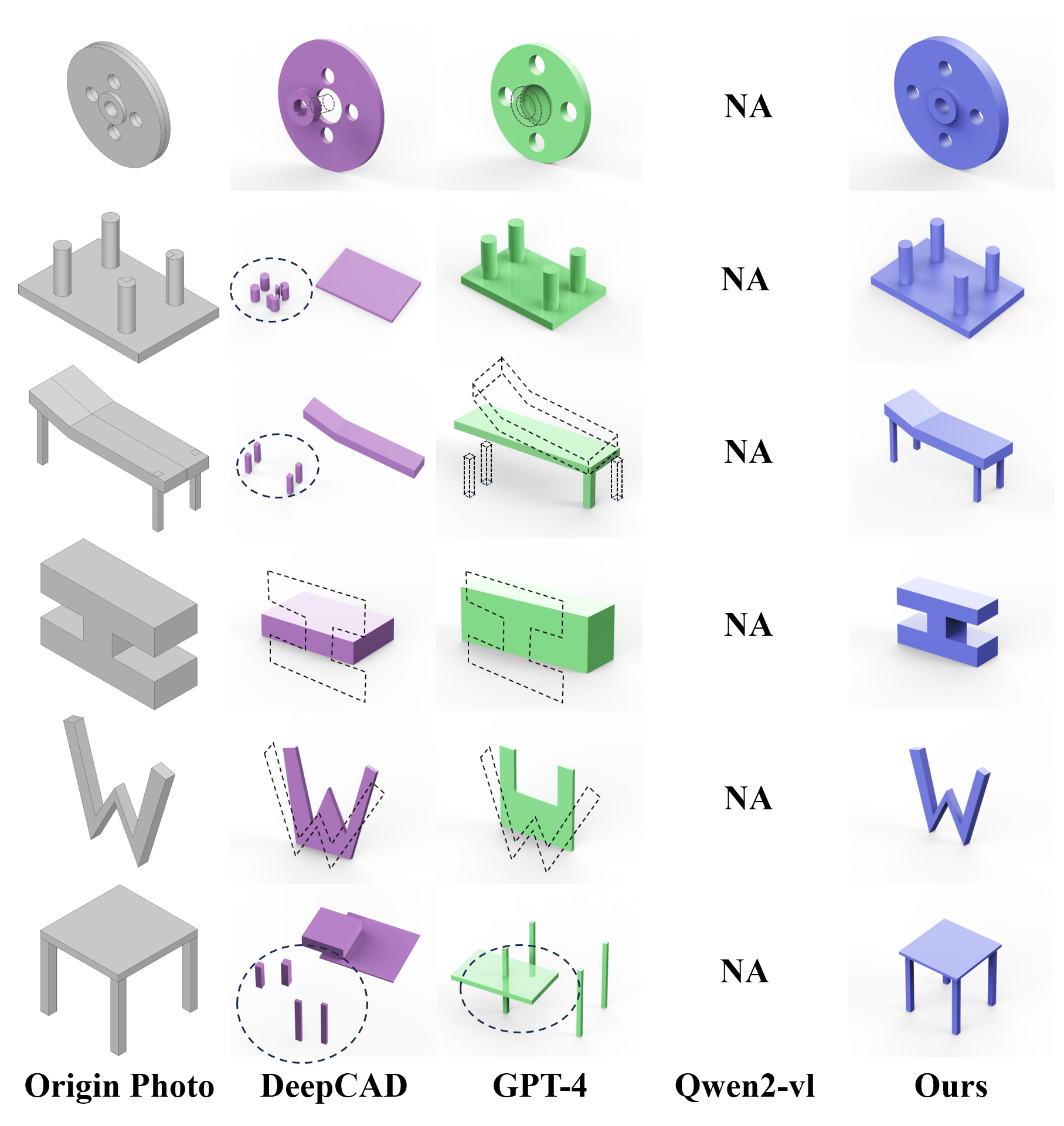}
    \caption{Comparison of different methods for image input scenarios}
    \label{fig:enter-label}
\end{figure}
\subsubsection{Qualitative Analysis}
In this section, we provide additional qualitative results on single-view image conditioning.As shown in Figure 3, we compared our approach against three representative methods. The first is DeepCAD, which exemplifies advanced generation techniques in CAD modeling. The second is GPT-4, representing the cutting-edge in closed-source multimodal large models. The third is Qwen2-VL-Max, one of the leading open-source multimodal large models. As observed, DeepCAD struggles with generating fine details, while GPT-4 exhibits limitations in spatial reasoning, frequently leading to errors in generated models. Qwen2-VL-Max, despite multiple attempts, consistently failed to render the generated JSON correctly. In contrast, our model produces outputs that are both accurate and aesthetically refined.
\subsubsection{Quantitative Comparison with Existing Methods}
In comparison with current CAD generation approaches, we use DeepCAD as a compare method. In addition, we compare our method with two recent autoregressive generative models, namely SkexGen \cite{xu2022skexgen} and HNC-CAD \cite{xu2023hierarchical}. We employ the same pre-trained visual encoder, ViT-L/14-336, and map its output to the same latent space for these methods. Additionally, we compare our method with the state-of-the-art multimodal large model GPT-4, with specific prompts detailed in the supplementary materials. As shown in Table 3, CAD-GPT achieves a median CD of 9.77, representing a 48\% reduction compared to the best-performing baseline HNC-CAD's 18.64. Furthermore, it achieves an 84\% lower than GPT-4's 62.64, demonstrating significantly higher reconstruction accuracy. In terms of the IR, CAD-GPT achieves a 91\% reduction compared to the best-performing baseline, HNC-CAD, and a 97\% reduction compared to the state-of-the-art multimodal model, GPT-4, demonstrating a significant improvement in generating valid CAD models. CAD-GPT also outperforms other methods on the two additional ACC metrics, demonstrating superior command generation accuracy. These results underscore CAD-GPT's superior precision and validity in CAD model reconstructions.

\begin{table}[ht]
    \centering
    \footnotesize
    \renewcommand{\arraystretch}{0.95}
    \begin{tabularx}{0.98\columnwidth}{lXXXXX}
        \toprule
        Model & IR $\downarrow$ & Median CD $\downarrow$ & ${ACC_{\text{cmd}}\uparrow}$ & ${ACC_{\text{param}}\uparrow}$ \\
        \midrule
        DeepCAD & 23.16 & 23.78 & 95.34 & 96.23 \\
        SkexGen & 22.32 & 20.45 & 95.82 & 96.63 \\
        HNC-CAD & 18.64 & 18.64 & 97.87 & 97.77 \\
        GPT-4 & 64.37 & 62.64 & 98.22 & 97.36 \\
        CAD-GPT & $\textbf{1.61}$ & $\textbf{9.77}$ & $\textbf{99.21}$ & $\textbf{98.87}$ \\
        \bottomrule
    \end{tabularx}
    \caption{Quantitative Evaluation of CAD Model Performance under Image Input Conditions}
    \label{tab:cad-comparison}
\end{table}
\begin{figure*}
    \centering
    \includegraphics[width=1\linewidth]{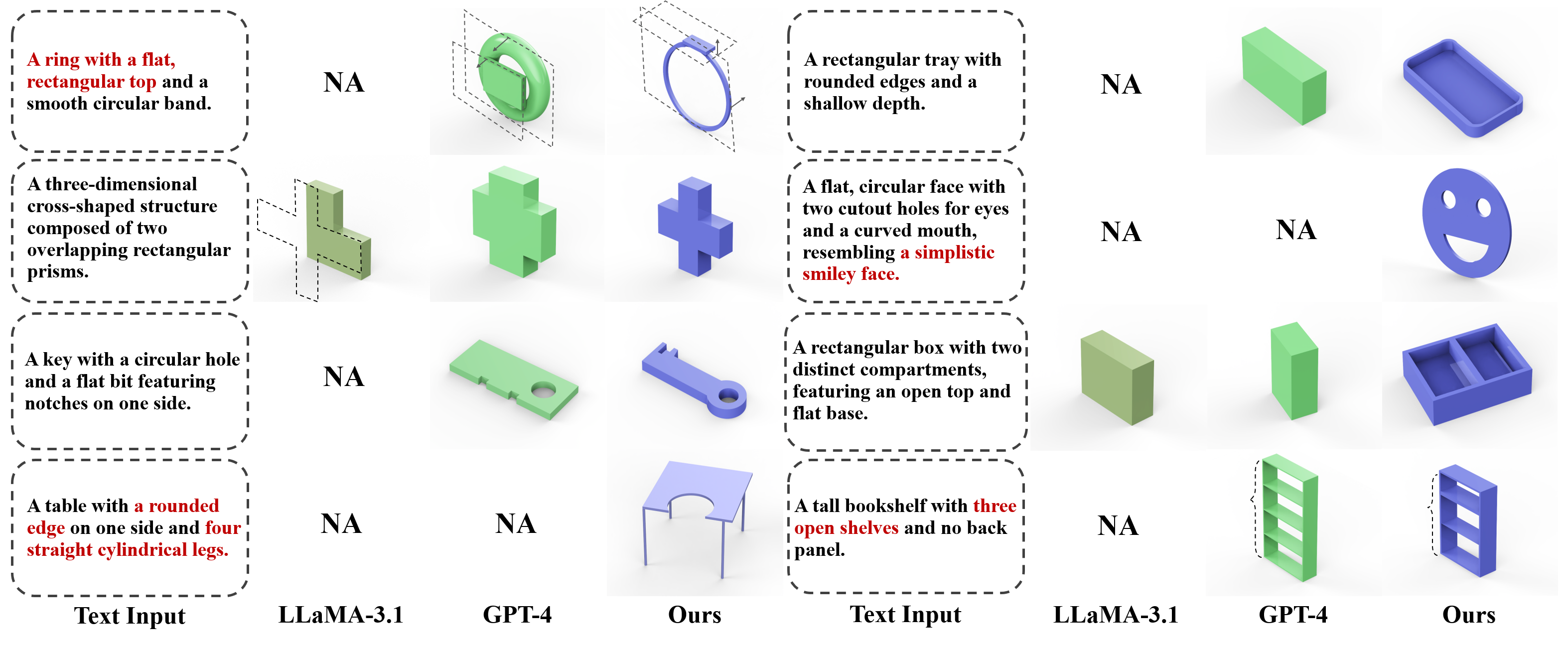}
    \caption{Comparison of different methods for text input scenarios}
    \label{fig:enter-label}
\end{figure*}

\subsection{CAD Generation from Text Descriptions}
\subsubsection{Qualitative Analysis}
In this section, we present additional qualitative results on text conditioning. Due to the lack of directly comparable CAD generation methods, we selected two representative large language models: GPT-4, a leading closed-source model, and LLaMA-3.1 (405B), a state-of-the-art open-source model. As illustrated in Figure 4, our model consistently generates high-precision, aesthetically pleasing outputs that align well with the textual descriptions across various scenarios. In contrast, GPT-4 frequently produces incorrect models with a high failure rate, while LLaMA-3.1 only occasionally succeeds in rendering models, and even then, the results often do not match the provided descriptions.
\subsubsection{Quantitative Comparison with Existing Methods}

We compare our approach with GPT-4 and the state-of-the-art open-source model LLaMA-3.1. We provide both models with the same background and input them with identical modeling instructions or text descriptions to generate the corresponding modeling code. As shown in Table 4, CAD-GPT achieves a median CD of 83\% lower than GPT-4's 187.52, and reduces the IR to 7.43, a 90\% decrease compared to GPT-4 and 92\% compared to LLaMA-3.1. This highlights CAD-GPT's superior accuracy and lower failure rates in CAD model reconstruction under text description inputs. In terms of ACC metrics, CAD-GPT outperforms the other two methods up to 6\%.
\begin{table}[ht]
    \centering
    \footnotesize
    \renewcommand{\arraystretch}{0.9}
    \begin{tabularx}{0.98\columnwidth}{lXXXX}
        \toprule
        Model & IR $\downarrow$ & Median CD$\downarrow$ &${ACC_{\text{cmd}}\uparrow}$ &${ACC_{\text{param}} \uparrow}$ \\
        \midrule
        LLaMA-3.1 & 98.68 & NA & NA & NA \\
        GPT-4 & 76.97 & 187.52 & 92.21 & 93.65 \\
        CAD-GPT & $\textbf{7.43}$ & $\textbf{28.33}$ & $\textbf{98.73}$ & $\textbf{98.12}$ \\
        \bottomrule
    \end{tabularx}
    \caption{Quantitative Evaluation of CAD Model Performance under Text Description Input Conditions}
    \label{tab:cad-comparison}
\end{table}

\subsection{Ablation Study}

The impact of the components proposed in CAD-GPT is evaluated in Table 5, focusing on CAD reconstruction metrics, including IR, mean CD, \( \text{ACC}_{\text{cmd}} \) and \( \text{ACC}_{\text{param}} \). The first row of the table shows the results when only the original data is trained, without our additional tokens and position embeddings. This configuration results in a decline in performance across CD distance, IR, as well as \( \text{ACC}_{\text{cmd}} \) and \( \text{ACC}_{\text{param}} \). The second row demonstrates the effects of incorporating only the three types of tokens. The incorporation of the three token series introduces 3D spatial positioning into the vocabulary, thereby enabling valid and accurate CAD reconstructions. The third row reports the results when both the three types of tokens and position embeddings are added. This configuration further reduces CD distance and IR while improving \( \text{ACC}_{\text{cmd}} \) and \( \text{ACC}_{\text{param}} \), demonstrating that our method effectively enhances modeling accuracy by mapping 3D spatial information into a one-dimensional space and constructing new learnable position encodings.

\begin{table}[ht]
    \centering   
    \small
    \footnotesize
    \renewcommand{\arraystretch}{0.9}
    \begin{tabularx}{0.98\columnwidth}{lXXXXX}
        \toprule
        Model & IR $\downarrow$ & Median CD$\downarrow$ &${ACC_{\text{cmd}}\uparrow}$ &${ACC_{\text{param}} \uparrow}$ \\        \midrule
        \textbf{Image Input} & & & \\
        w/o Loc & 37.15 & 161.31 & 90.45 & 91.37 \\
        w/o Emb & 4.31 & 27.98 & 91.63 & 91.55 \\
        CAD-GPT & \textbf{1.61} & \textbf{9.77} & \textbf{99.21} & \textbf{98.87} \\
        \midrule
        \textbf{Text Input} & & & \\
        w/o Loc & 40.23 & 145.87 & 83.42 & 83.44 \\
        w/o Emb & 10.12 & 29.58 & 87.54 & 88.23 \\
        CAD-GPT & $\textbf{7.43}$ & $\textbf{28.33}$ & $\textbf{98.73}$ & $\textbf{98.12}$ \\
        \bottomrule
    \end{tabularx}
    \caption{Ablation Study with Image and Text as Input}
    \label{tab:cad-comparison}
\end{table}
\section{Conclusion}
In this paper, we introduce CAD-GPT, a multimodal large model enhanced with the 3D Modeling Spatial Localization Mechanism to improve spatial reasoning capabilities. Our model excels at inferring variations in sketch orientations, changes in 3D spatial positions, and accurately rendering 2D sketches. Leveraging these capabilities, CAD-GPT demonstrates exceptional performance in generating precise CAD models under both image and text input conditions.

\section{Acknowledgments}
The authors would like to express their gratitude to Guojun Yin for his support in this work. He is currently an MLLM Algorithm senior research at Meituan. We also wish to thank Rundi Wu, the author of DeepCAD, for his invaluable guidance and assistance throughout the development of this paper. This work was supported by the National Natural Science Foundation of China (No. 92167205, 62025305, 61933009, 62432009, 62422311, U22A2050), and the Shanghai Committee of Science and Technology, China (No. 24TS1413500).

\bigskip

\bibliography{aaai25}

\end{document}